\documentclass{article}
\usepackage{spconf,amsmath,graphicx,hyperref}
\usepackage{multirow}
\usepackage{cite}
\usepackage{amssymb,amsfonts}
\usepackage{algorithmic}
\usepackage{textcomp}
\usepackage{xcolor}
\usepackage{enumitem}

\title{Lamer-SSL: Layer-aware mixture of LoRA experts for continual multilingual expansion of self-supervised models without forgetting}
\name{Jing Xu, Minglin Wu, Xueyuan Chen, Xixin Wu, Helen Meng}
\address{The Chinese University of Hong Kong}
\begin{document}
\ninept
\maketitle
\begin{abstract}
Despite their impressive performance, self-supervised speech models often struggle to generalize to new languages and tend to forget previously acquired knowledge during continual training. To address this, we propose Lamer-SSL, a parameter-efficient framework that integrates a \textbf{L}ayer-\textbf{A}ware \textbf{M}ixtur\textbf{E} of Lo\textbf{R}A Experts (Lamer) module with a replay strategy. The Lamer module enables flexible balancing between shared and language-specific representations, while layer-aware expert allocation assigns more experts to deeper layers where semantic information is richer. Meanwhile, the replay strategy retains prior knowledge using minimal data, mitigating forgetting during continual training. 
Experiments on automatic speech recognition (ASR) and language identification (LID) demonstrate that Lamer-SSL extends self-supervised models to new languages effectively while maintaining strong performance on previously learned languages with only 2.14\% parameters being trainable.
\end{abstract}
\section{Introduction}
\label{sec:intro}

Self-supervised learning (SSL) has greatly advanced speech processing by enabling models to learn universal speech representations from large-scale unlabeled audio. Models such as Wav2vec 2.0\cite{wav2vec}, HuBERT\cite{hubert}, and WavLM\cite{wavlm} have shown strong performance across a range of downstream tasks including automatic speech recognition (ASR), speech synthesis, and speaker verification. 

To extend SSL to multilingual scenarios, recent work has primarily relied on large-scale joint training. Models such as XLSR~\cite{xlsr}, mHuBERT~\cite{mhubert}, mHuBERT-147~\cite{mhubert147}, and MMS\cite{scalingmms} are trained on dozens of languages simultaneously, often employing sampling-based strategies to mitigate data imbalance. 
Typical approaches include upsampling low-resource languages~\cite{xlsr} and temperature-based sampling~\cite{xlmr}, which increase the relative contribution of low-resource languages. 
While effective, they suffer from critical drawbacks. First, training from scratch is prohibitively expensive, especially as new languages and dialects continuously emerge. Second, static sampling fails to resolve language interference, where high-resource languages suppress low-resource ones. Moreover, the brute-force upsampling strategy introduces redundancy, increases training time, and lacks scalability, since adding new languages typically requires reprocessing the entire dataset and retraining the model from scratch. These challenges underscore the need for continual multilingual training, where models can be incrementally extended to new languages.

A major obstacle in continual training is catastrophic forgetting, where adapting to new languages often degrades performance on previously learned ones. Parameter-efficient adaptation methods such as LoRA~\cite{lora}, DoRA~\cite{dora}, adapters~\cite{ssladaptive}, and prefix-tuning~\cite{prefixtuning} mitigate this by freezing the backbone and injecting lightweight modules. For instance, LoRA has been shown to extend HuBERT to Mandarin with reduced forgetting~\cite{seamlesslanguage}. However, while such methods help retain prior knowledge, they often struggle with language interference, limiting their effectiveness in multilingual scenarios.

Meanwhile, Mixture-of-Experts (MoE) has become a powerful approach for balancing shared and task-specific knowledge. By activating only a subset of experts, MoE scales model capacity without proportional computational cost~\cite{firstmoe,switchformer,gshard,moe}. In natural language processing, MoE has been widely explored for multilingual and multi-task learning. In speech, recent studies\cite{speechconformermoe,mole,speechinformed} show that expert specialization improves multilingual speech recognition and representation learning. 
However, most designs allocate experts uniformly across layers and are typically trained from scratch, overlooking the hierarchical nature of SSL representations, where shallow layers capture acoustic features and deeper layers encode more semantic and linguistic information~\cite{analysis,analysis2,analysis3}. 
This mismatch limits the effectiveness and scalability of MoE in continual multilingual adaptation.

To address these limitations, we propose Lamer-SSL, a parameter-efficient continual training framework that integrates a layer-aware Mixture of LoRA Experts (Lamer) module with a replay strategy. Lamer combines the efficiency of LoRA with the flexibility of MoE, enabling dynamic sharing across languages while retaining language-specific capacity. Unlike prior work, our layer-aware expert allocation assigns more experts to deeper layers, aligning with the hierarchical nature of SSL representations. In addition, the replay strategy revisits minimal prior data to further mitigate forgetting without requiring access to the entire historical dataset. Our main contributions are as follows:
\begin{itemize}[noitemsep, topsep=1pt, parsep=1pt, partopsep=1pt]
\item We propose Lamer-SSL, a parameter-efficient continual training framework for extending SSL models to new languages.
\item We combine the efficiency of LoRA with the flexibility of MoE to enable scalable multilingual extension.
\item We design a layer-aware expert allocation strategy that assigns experts based on the hierarchical nature of SSL models.

\item Experiments on ML-SUPERB monolingual ASR and language identification tasks confirm that Lamer-SSL effectively extends models to new languages while preserving prior capabilities, using only 2.14\% trainable parameters.
\end{itemize}
\begin{figure*}
\centering
\begin{minipage}[b]{0.2\linewidth}
  \centering
  \centerline{\includegraphics[width=2in]{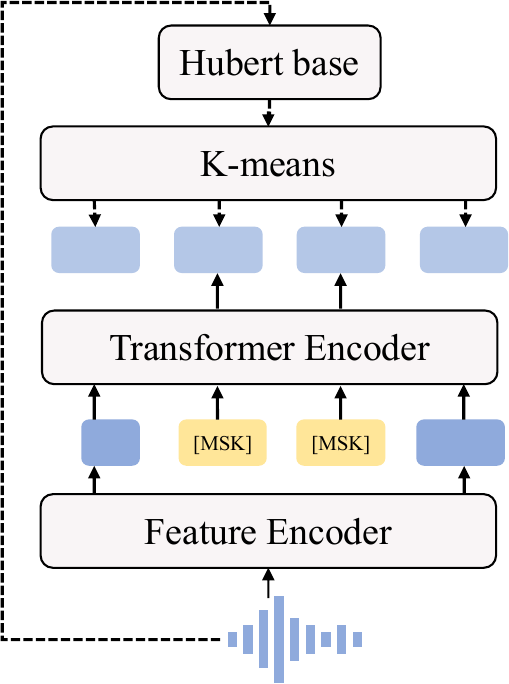}}
  \centerline{(a)}\medskip
\end{minipage}
\hspace{0.12\linewidth}
\begin{minipage}[b]{0.2\linewidth}
  \centering
  \centerline{\includegraphics[width=2in]{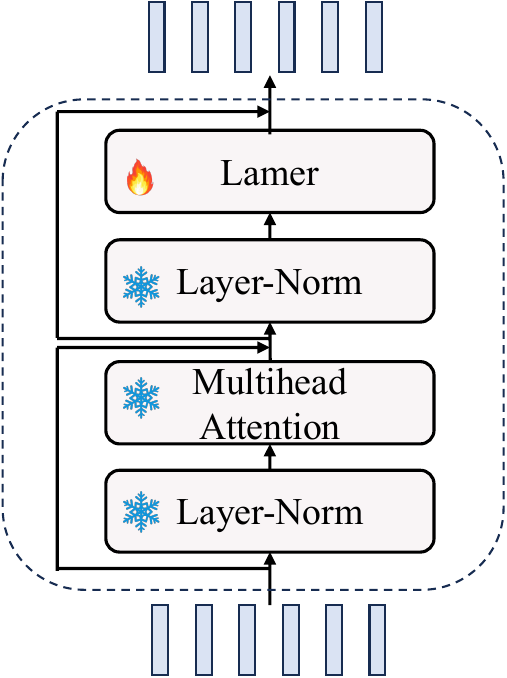}}
  \centerline{(b)}\medskip
\end{minipage}
\hspace{0.12\linewidth}
\begin{minipage}[b]{0.2\linewidth}
  \centering
  \centerline{\includegraphics[width=2in]{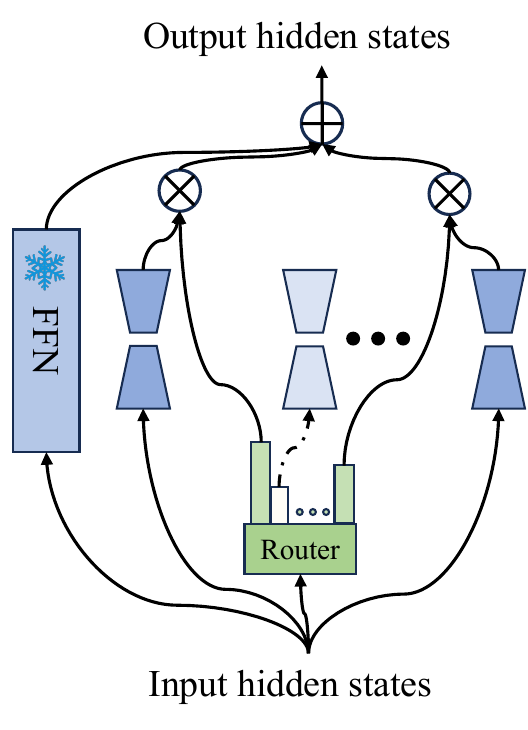}}
  \centerline{(c)}\medskip
\end{minipage}
%
\vspace{-1em}
\caption{Overview of Lamer-SSL. (a) Architecture of HuBERT-based SSL models. (b) Transformer block with Lamer module. (c) Architecture of Lamer module, the router selects the Top-K experts based on the input. Only the LoRA experts and the router are trainable during training.}

\label{fig:lamer-ssl}
\end{figure*}
\section{Methodology}
\label{sec:method}
Lamer-SSL is a parameter-efficient framework designed for continual multilingual extension of SSL models. It integrates a Mixture of LoRA Experts with a layer-aware allocation strategy to support scalable language extension while preserving prior knowledge. This section introduces the architecture of Lamer-SSL (Section \ref{sec:Lamer}), the layer-aware expert allocation strategy (Section \ref{sec:layer-aware allocation}), the replay strategy for mitigating forgetting (Section \ref{sec:replay strategy}), and the training objective used during continual learning (Section \ref{sec:training objective}).

\subsection{Architecture of Lamer-SSL}
\label{sec:Lamer}
As illustrated in Figure~\ref{fig:lamer-ssl}(a), HuBERT-based SSL models typically consist of a convolutional feature encoder followed by a Transformer encoder with $L$ blocks. Each Transformer block contains multi-head self-attention and a feed-forward network (FFN). In Lamer-SSL, the original FFN is replaced with a Lamer module, which includes a frozen FFN backbone, a set of LoRA experts, and a router. 

LoRA extends model capacity by injecting trainable low-rank matrices into the FFN's projection layers, while keeping the original weights $W_0\in \mathbb{R}^{d_{out}\times d_{in}}$ frozen. Prior work \cite{seamlesslanguage} shows that LoRA adapters can extend SSL models to new languages with reduced forgetting. However, a single LoRA module lacks sufficient expressiveness for multilingual modeling. Lamer-SSL addresses this by introducing a Mixture of LoRA Experts mechanism, where each expert specializes in distinct linguistic patterns. 

Given input hidden states $h_0$, the frozen FFN produces a base representation:
\begin{align}
h_{base} = W_0h_0
\end{align}
Each expert $E_k(h_0)$ applies a low-rank transformation:
\begin{align}
E_k(h_0)= B_kA_kh_0
\end{align}
with $A_k \in \mathbb{R}^{r\times d_{in}}$, $B_k \in \mathbb{R}^{d_{out}\times r}$, and $r\ll min(d_{in},d_{out})$. A router computes the routing probabilities over experts:
\begin{align}
p = softmax(W_rh_0)
\end{align}
A Top-K operator selects the K largest entries of $p$ and re-normalizes their weights: 
\begin{align}
R(h_0)_k =
    \begin{cases}
    \frac{p_k}{\sum_{j \in \mathcal{T}(h_0)} p_j}, & \text{if } k \in \mathcal{T}(h_0), \\
    0, & \text{otherwise},
    \end{cases}
\end{align}
where $\mathcal{T}(h_0)$ denotes the set of Top-K experts. 
The final output is:

 \begin{align}
h_{\text{out}} = h_{base} + \sum_{k=1}^{N} R(h_0)_k \cdot E_k(h_0)
\end{align}
This sparse routing ensures that each token is processed by the most relevant experts, enabling both language-specific adaptation and cross-lingual generalization without requiring explicit language identifiers. 
Gradients are only backpropagated through the selected experts, allowing for efficient scaling to new languages.

\subsection{Layer-aware expert allocation}
\label{sec:layer-aware allocation}
Conventional MoE architectures often allocate experts uniformly across layers, assuming all layers require the same degree of specialization. However, analyses of SSL models reveal that shallow layers mainly capture low-level acoustic features such as speaker, pitch, and energy, while deeper layers encode higher-level semantic and linguistic information.

To align expert capacity with representational complexity, Lamer-SSL adopts a layer-aware expert allocation strategy. Early layers, which capture mainly language-independent acoustic information, require fewer experts. In contrast, deeper layers encode more language-dependent semantic information, thus necessitating more experts to enable effective specialization. This hierarchical allocation improves parameter efficiency and concentrates modeling capacity where it is most beneficial for multilingual generalization.
\subsection{Replay Strategy}
\label{sec:replay strategy}
Beyond architectural design, Lamer-SSL incorporates a replay strategy to further mitigate forgetting. During continual training, a small subset of samples from previously learned languages is interleaved with data from new languages. These replay samples are randomly selected and do not require access to the full original training dataset. By revisiting prior language distributions, the model maintains prior knowledge without reliance on large historical corpora, thereby promoting scalable and stable multilingual extension.

\subsection{Training Objective}
\label{sec:training objective}

The training objective of Lamer-SSL extends HuBERT's standard masked prediction loss with an auxiliary load balancing loss to ensure effective representation learning and stable expert utilization.  

Following HuBERT, for each masked frame at time step $t$, the model predicts the cluster index $z_t$ obtained via K-means clustering over SSL representations:
\begin{align}
\mathcal{L}_{\text{mask}} = - \sum_{t \in \mathcal{M}} \log P(z_t \mid \hat{h}_t),
\end{align}
where $\mathcal{M}$ denotes the set of masked time steps, and $\hat{h}_t$ is the contextualized representation. This loss is jointly applied to both new language data and replayed samples, allowing the model to integrate novel linguistic features without sacrificing prior capabilities.

To prevent expert collapse, a load balancing loss is introduced. Without this constraint, the router may assign most tokens to only a few experts, leaving others underutilized. Let $m_{k}$ denote the average routing probability for expert $k$, and $f_{k}$ the fraction of tokens actually dispatched to expert $k$. The load balancing loss is:
\begin{align}
\mathcal{L}_{\text{lb}} = N \cdot \sum_{k=1}^{N} m_{k} \cdot f_{k},
\end{align}
where $N$ is the number of experts. This term encourages consistency between routing probabilities and actual token assignments, thereby promoting more balanced expert utilization.

The final training objective is a weighted combination:
\begin{align}
\mathcal{L} = \mathcal{L}_{\text{mask}} + \lambda \mathcal{L}_{\text{lb}},
\end{align}
where $\lambda$ controls the trade-off between prediction accuracy and expert balance. This composite objective enables Lamer-SSL to extend language coverage efficiently, while maintaining stable expert specialization and mitigating forgetting.

\section{Experimental Setup}
\label{sec:setup}

\subsection{Datasets}
We follow mHuBERT-147~\cite{mhubert147} for data selection, using its public manifests\footnote{\url{https://huggingface.co/utter-project/mHuBERT-147-base-3rd-iter/tree/main/manifest}} to extract Mandarin (zh-CN) and Cantonese (zh-HK, yue) samples from CommonVoice 11.0~\cite{commonvoice}. 
For Mandarin, we additionally include Thchs-30~\cite{thchs30}, AISHELL-1~\cite{aishell}, and AISHELL-3~\cite{aishell3}, consistent with mHuBERT-147, but exclude VoxLingua107, resulting in a slightly smaller Mandarin corpus. 
For replay, 100 hours of English speech are randomly sampled from CommonVoice 11.0. 
No data up-sampling is applied and utterances outside 2$\sim$30 second range are filtered. 
The final dataset comprises 105.072 hours of Cantonese (same as mHuBERT-147), 341.867 hours of Mandarin (slightly less than 382.7 hours in mHuBERT-147), and 100.555 hours of English (compared to 47,210.7 hours in mHuBERT-147).  
This setting enables us to assess Lamer-SSL’s ability to incorporate new languages while preserving prior knowledge. 

For evaluation, we target two downstream tasks: (1) monolingual ASR on both CommonVoice (in-domain) and Fleurs~\cite{fleurs} (out-of-domain), and (2) language identification (LID) using 1-hour mixtures of Mandarin, Cantonese and English from both datasets.  

\subsection{Training Configurations}
We continually train HuBERT-Large to improve its Mandarin and Cantonese performance. Training targets are obtained via K-means clustering on the 9th-layer features of mHuBERT-147. 
For each language, 50 hours of speech are randomly selected for clustering, using MiniBatchKMeans with K-means++ initialization, a batch size of 10k frames, and 20 random seeds.  

The model is trained for 400k steps with a peak learning rate of $1.5 \times 10^{-3}$. Experts are allocated in groups of six Transformer layers, with each group assigned the same number of experts in a progressive configuration of 2, 4, 6, 8 from shallow to deep layers. Each expert is a rank-12 LoRA module. The router selects the Top-2 experts per token, resulting in 2.14\% of parameters being trainable. The load-balancing coefficient $\lambda$ is set to $0.001$.
\begin{table*}
\centering
\caption{Performance comparison across ASR and LID tasks on three languages. Bold marks the best, "avg" denotes the average.}
\label{tab: main results}
\resizebox{0.95\linewidth}{!}{
\begin{tabular}{lcccccccccccc}
\hline
\multirow{3}{*}{System}   & \multicolumn{8}{c}{Monolingual ASR (CER$\downarrow$)} & \multicolumn{4}{c}{\multirow{2}{*}{LID (ACC$\uparrow$)}} \\
\cline{2-9}
& \multicolumn{4}{c}{CommonVoice}    & \multicolumn{4}{c}{Fleurs}  \\ 
\cline{2-13}
&eng & cmn & yue & avg & eng &cmn &yue & avg & eng &cmn & yue & avg\\
\hline\hline
mHuBERT-147 (1iter) &30.2 & 24.7 & 21.8 & 25.57 & 26.6 & 25.6 & 25.3 & 25.83 & 93.41 & 91.30 &93.10 &92.60\\
mHuBERT-147 (2iter)   & 21.2 & 17.4 & 15.8 & 18.13 & 18.2 & 16.4 & 17.2 & 17.27 & 97.60 & 96.89 & 98.85& 97.78\\
mHuBERT-147 (3iter)    &18.5 & 15.5 & 14.8 & 16.27 & 15.9 &15.1 & 15.6 &15.53 & 98.20 & 96.89 & 96.55 & 97.21\\
HuBERT-Large  & 11.5 & 21.2 & 17.6& 16.77 & 10.4 & 21.2 & 18.1  & 16.57 & 97.60 & 90.68 & 96.55 & 94.94\\
HuBERT-LORA & 11.0 & 11.3 & 11.1 &11.13 & 10.0 & 10.7 &11.9 &10.87 & 98.20 & 98.76 & \textbf{99.43} & 98.80\\
 \hline

Lamer-SSL (Ours) & \textbf{9.9} & \textbf{10.8} &\textbf{10.8} &\textbf{10.50} & \textbf{9.1} & \textbf{9.9} & \textbf{11.4} & \textbf{10.13} & \textbf{100.00} & \textbf{99.38} & 98.28 & \textbf{99.22} \\

 \hline
\end{tabular}
}
\end{table*}
\subsection{Evaluation Configurations}
Evaluation follows ML-SUPERB protocols for monolingual ASR and LID tasks under the 1-hour fine-tuning scenario. ASR performance is reported with Character Error Rate (CER), while LID is evaluated by Accuracy (ACC). All experiments follow the official ESPnet recipes\footnote{\url{https://github.com/espnet/espnet/tree/master/egs2/ml_superb/asr1}}, with a fine-tuning learning rate of $1\times10^{-4}$. For simplicity in the following, the language codes cmn, yue, and eng are used to refer to Mandarin, Cantonese, and English, respectively.

To stabilize low-resource training, we use intermediate representations as the input: the 22nd layer of HuBERT-Large and 11th layer of HuBERT-Base for ASR task, and the 21st layer of HuBERT-Large and the 10th layer of HuBERT-Base for LID task.

\section{Experimental results}
\label{sec:results}
\subsection{Comparing systems}
We compare Lamer-SSL against three representative systems:
\begin{itemize}
\item \textbf{HuBERT-Large}: An English-only SSL model pretrained on 60k hours of Libri-Light, used as the base for all continual training experiments.

\item \textbf{mHuBERT-147 (iter 1-3)}: A multilingual HuBERT model trained on 90k hours speech from 147 languages. We include its three intermediate checkpoints to track performance gains and benchmark the effect of multilingual pretraining.

\item \textbf{HuBERT-LoRA}: A continual training baseline where LoRA modules with rank 24 are directly injected into HuBERT-Large without expert routing or layer-aware allocation. This isolates the effect of low-rank adaptation.

\end{itemize}
\subsection{Main results}

Table~\ref{tab: main results} demonstrates that Lamer-SSL achieves the best overall performance across both ASR and LID tasks. On ASR, it records average CERs of 10.50\% (CommonVoice) and 10.13\% (Fleurs), outperforming HuBERT-LoRA by 0.63\% and 0.74\%, respectively. These gains highlight the benefits of modular expert routing and layer-aware allocation beyond simple low-rank adaptation.

Despite mHuBERT-147 undergoing extensive multilingual pretraining, Lamer-SSL outperforms it with far fewer resources and no data upsampling. For instance, Lamer-SSL reduces CER from 16.27\% to 10.50\% on CommonVoice and from 15.53\% to 10.13\% on Fleurs, despite starting from an English-only model.

For LID, Lamer-SSL also sets a new state-of-the-art, achieving 99.22\% average accuracy across English, Mandarin, and Cantonese. This surpasses all other systems, including HuBERT-LORA (98.80\%) and mHuBERT-147 (3iter) (97.21\%), indicating strong language discrimination and minimal forgetting.

Overall, these results validate the effectiveness of Lamer-SSL in extending SSL models to new languages while maintaining high performance and parameter efficiency. Its modular design enables robust adaptation without the need to reprocess the entire dataset or undergo costly multilingual pretraining and large-scale retraining.
\subsection{Analysis of expert activation patterns}
\begin{figure}[ht]
\centering
\vspace{-1.4em}
\includegraphics[width=0.94\linewidth]{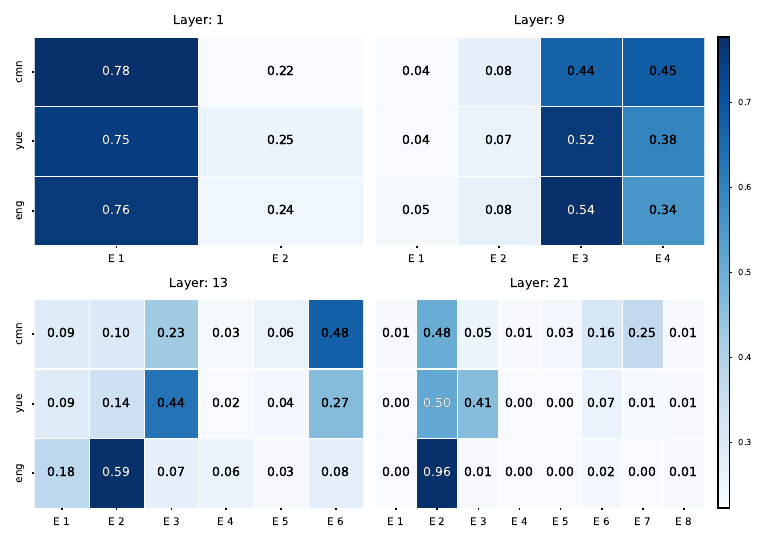}
%
\vspace{-1em}
\caption{Expert activation weights across languages at four layers.}
\label{fig:weights_patterns}
 \vspace{-1em}
\end{figure}
Figure~\ref{fig:weights_patterns} visualizes expert activation weights for English, Cantonese, and Mandarin across four Transformer layers (1, 9, 13, 21), with darker shades indicating stronger activation.

In shallow layers (\textit{e.g.,} Layer 1), all languages exhibit similar expert activation probabilities, indicating shared acoustic characteristics across languages. As depth increases, activation becomes more language-specific. At Layer 21, all three languages activate Expert 2 with distinct activations and diverge in secondary expert selections: English favors Expert 6, Mandarin Expert 7, and Cantonese Expert 3, revealing clear specialization. This progression underscores how deeper layers facilitate semantic specialization, supporting the effectiveness of Lamer-SSL’s layer-aware expert allocation.
\subsection{Ablation study}
To assess the contribution of each component, we conduct an ablation study by removing the layer-aware expert allocation, load balancing loss, and replay strategy. All variants are trained under identical settings with the same LoRA  and routing configurations.

As shown in Table~\ref{tab: ablationstudy}, each removal leads to a noticeable degradation in ASR performance. Notably, excluding the replay strategy causes severe degradation on English, confirming its role in mitigating forgetting. These results underscore that all three components are essential for stable and effective multilingual extension.
\begin{table}[ht!]
\centering
\vspace{-1em}
\caption{Ablation study}
\label{tab: ablationstudy}
\resizebox{\linewidth}{!}{
\begin{tabular}{lcccccccccccc}
\hline
\multirow{3}{*}{System}   & \multicolumn{8}{c}{Monolingual ASR (CER$\downarrow$)}\\
\cline{2-9}
& \multicolumn{4}{c}{CommonVoice}    & \multicolumn{4}{c}{Fleurs}  \\ 
\cline{2-9}
&eng & cmn & yue & avg & eng &cmn &yue & avg \\
\hline

Lamer-SSL (Ours)& \textbf{9.9} & \textbf{10.8} &\textbf{10.8} &\textbf{10.50} & \textbf{9.1} & \textbf{9.9} & \textbf{11.4} &\textbf{10.13}  \\

-Layer-aware allocation & 10.5 & 11.6 &11.4 &11.17 & 9.4 & 11.4 & 11.6 & 10.80 \\
-Load balancing loss &  10.8& 12.1 &12.1 &11.67 & 9.6 & 12.2 & 13.0&11.60\\ 
-Replay strategy & 26.9 & 11.1 &10.9 &16.30 &23.4&10.6 &11.6 &15.20\\

\hline
\end{tabular}
}
\vspace{-2em}
\end{table}

\subsection{Affects of expert allocation strategies}
We further investigate how expert distribution affects performance by testing five allocation schemes with the same total number of experts, replay strategy and load-balancing regularization. Table~\ref{tab: differenct expert allocation} reports the ASR performance on CommonVoice and Fleurs datasets.

Our default configuration (Lamer-SSL (Ours)), which allocates experts in a progressively increasing pattern (2/4/6/8), achieves the best overall results, with the lowest average CER of 10.50\% on CommonVoice and 10.13\% on Fleurs. While other schemes also allocate more experts to deeper layers, the progressively increasing configuration (2/4/6/8) consistently outperforms the imbalanced scheme (2/2/8/8), suggesting that gradual expert scaling across depth better supports multilingual generalization.
\begin{table}[ht!]
\centering
\vspace{-1em}
\caption{Comparison under different expert allocation strategies}
\label{tab: differenct expert allocation}
\resizebox{\linewidth}{!}{
\begin{tabular}{lcccccccccccc}
\hline
\multirow{3}{*}{System}   & \multicolumn{8}{c}{Monolingual ASR (CER$\downarrow$)}\\
\cline{2-9}
& \multicolumn{4}{c}{CommonVoice}    & \multicolumn{4}{c}{Fleurs}  \\ 
\cline{2-9}
&eng & cmn & yue & avg & eng &cmn &yue & avg \\
\hline
Lamer-SSL (Ours) & \textbf{9.9} & \textbf{10.8} &\textbf{10.8} &\textbf{10.50} & \textbf{9.1} & \textbf{9.9} & \textbf{11.4} &\textbf{10.13}\\
Lamer-SSL-(8/6/4/2) & 10.7& 11.2 &11.3 & 11.07 & 9.4 &10.7 &12.3 & 10.80  \\
Lamer-SSL-(8/2/2/8) &10.5 & 11.5 &11.3 & 11.10 & 9.4 &11.3 &12.0 &10.90 \\ 
Lamer-SSL-(2/8/8/2) & 10.7 & 11.2 &10.9 &10.93 & 9.5 & 10.5 & 11.4 & 10.47 \\
Lamer-SSL-(2/2/8/8) & 10.9 & 11.8 & 11.4 & 11.37 & 9.7 & 10.8 & 12.2 & 10.90  \\
\hline
\end{tabular}
}
\vspace{-1em}
\end{table}
\section{Conclusions}
\label{sec:conclusion}

We present Lamer-SSL, a modular framework for continual multilingual self-supervised learning. By combining Mixture-of-LoRA Experts with layer-aware expert allocation and a replay-based strategy, Lamer-SSL achieves strong ASR and LID performance efficiently. Experiments show clear gains over both multilingual pretraining and LoRA-only baselines. Analysis of expert activations further confirms that deeper layers support language-specific routing, validating our design. In the future, we aim to develop more adaptive expert allocation mechanisms and extend Lamer-SSL to more languages.

\section{Acknowledgement}
This study was supported by Human-Computer Communications Laboratory  (HCCL), Department of Systems Engineering and Engineering Management, The Chinese University of Hong Kong, Hong Kong SAR, China and the Centre for Perceptual and Interactive Intelligence (CPII) Ltd., a CUHK-led InnoCentre under the InnoHK initiative of the Innovation and Technology Commission of the Hong Kong Special Administrative Region Government.

\bibliographystyle{IEEEbib}
\bibliography{strings,refs}

\end{document}